\documentclass[conference]{IEEEtran}
\IEEEoverridecommandlockouts
\usepackage{amsmath,amssymb,amsfonts}
\usepackage{algorithmic}
\usepackage{graphicx}
\usepackage{textcomp}
\usepackage{xcolor}
\def\BibTeX{{\rm B\kern-.05em{\sc i\kern-.025em b}\kern-.08em
    T\kern-.1667em\lower.7ex\hbox{E}\kern-.125emX}}

\usepackage{xspace}
\usepackage{empheq}
\usepackage{algorithm}
\usepackage{subfigure}
\usepackage{multirow}
\usepackage[pagebackref,breaklinks,colorlinks,citecolor=blue,linkcolor=blue,bookmarks=false]{hyperref}
\usepackage{caption}
\usepackage{balance}
\usepackage{makecell}

\DeclareMathOperator*{\argmin}{arg\,min}

\def\onedot{.\@\xspace}
\def\eg{\emph{e.g}\onedot} 
\def\ie{\emph{i.e}\onedot}

\def\etal{\emph{et al}\onedot}
\def\aka{\emph{a.k.a}\onedot}

\newcommand{\Eref}[1]{Eq.~(\ref{#1})}
\newcommand{\Fref}[1]{Fig.~\ref{#1}}
\newcommand{\Tref}[1]{Table~\ref{#1}}
\newcommand{\Aref}[1]{Algorithm~\ref{#1}}

\newcommand{\calD}{{\mathcal{D}}}
\newcommand{\calF}{{\mathcal{F}}}
\newcommand{\calL}{{\mathcal{L}}}
\newcommand{\calW}{{\mathcal{W}}}
\newcommand{\calX}{{\mathcal{X}}}

\newcommand{\norm}[1]{\| #1 \|}

\DeclareRobustCommand*{\IEEEauthorrefmark}[1]{%
  \raisebox{0pt}[0pt][0pt]{\textsuperscript{\footnotesize #1}}%
}

\begin{document}

\title{
Unsupervised Deep One-Class Classification with \\ Adaptive Threshold based on Training Dynamics
}

\author{\IEEEauthorblockN{
Minkyung Kim\IEEEauthorrefmark{1},
Junsik Kim\IEEEauthorrefmark{2}, 
Jongmin Yu\IEEEauthorrefmark{3} and
Jun Kyun Choi\IEEEauthorrefmark{1}}
\IEEEauthorblockA{
\IEEEauthorrefmark{1}\textit{School of Electrical Engineering}, \textit{KAIST}, Republic of Korea \\
\IEEEauthorrefmark{2}\textit{School of Engineering and Applied Sciences}, \textit{Harvard University}, U.S.A. \\
\IEEEauthorrefmark{3}\textit{Department of Engineering}, \textit{King's College London}, United Kingdom\\
mkkim1778@kaist.ac.kr,
mibastro@gmail.com,
jm.andrew.yu@gmail.com,
jkchoi59@kaist.edu}
}

\maketitle

\begin{abstract}
One-class classification has been a prevailing method in building deep anomaly detection models 
under the assumption that a dataset consisting of normal samples is available.
In practice, however, abnormal samples are often mixed in a training dataset, 
and they detrimentally affect the training of deep models, which limits their applicability. 
For robust normality learning of deep practical models, we propose an unsupervised deep one-class classification that learns normality from pseudo-labeled normal samples, \ie, outlier detection in single cluster scenarios.
To this end, we propose a pseudo-labeling method by an adaptive threshold selected by ranking-based training dynamics.
The experiments on 10 anomaly detection benchmarks show that our method effectively improves performance on anomaly detection by sizable margins.
\end{abstract}


\section{Introduction}
Anomaly detection (AD) is widely utilized to identify observations that deviate significantly from what is considered normal in each application domain~\cite{chandola2009anomaly}.
By leveraging recent successes in deep learning, deep learning-based AD has been actively researched and has shown high capabilities across various applications
, such as 
intrusion detection~\cite{yu2021unusual}, 
fraud detection~\cite{zhang2021fraudre},
defect detection~\cite{bergmann2019mvtec, wangzhe2019approaches}, and 
surveillance system~\cite{singh2018eye}.
In training deep AD models, normal samples are considered far easier obtainable than abnormal samples, which supports the common assumption that all the training data are normal samples~\cite{oza2018one, ruff2018deep, perera2019ocgan, wu2019deep, gong2019memorizing, bergman2020classification, goyal2020drocc}.
Then, the aim of model training is to describe normality accurately~\cite{ruff2018deep}.
An abnormal sample is defined as an observation deviating considerably from the learned normality.
In the view of learning normality from normal samples, 
one-class classification (OCC) has been a prevailing method in building deep AD models~\cite{ruff2021unifying}.

OCC-based deep AD models define the concept of normality as a discriminative decision boundary through mapping normal data to a compact representation.
However, in practice, abnormal samples are often mixed in a training dataset due to the nature of real data distributions that include anomalous tails.
Identifying labels for all the data to train a deep one-class classifier is an expensive process in terms of the time and effort of domain experts.
This requirement for normal samples limits the applicability of deep models.
Furthermore, abnormal samples mixed in a training dataset have a detrimental effect on the normality that a model learns~\cite{zong2018deep, beggel2019robust}.
Thus, a practical deep AD model needs to be robust in learning normality with a mixed dataset of normal and abnormal samples without labels, \ie, unsupervised AD.

While deep AD with a training dataset consisting of only normal samples, \aka, semi-supervised AD, has been extensively studied, research dealing with unlabeled datasets, \aka, unsupervised AD, has been less explored.
Unsupervised AD can be grouped into two categories: robust methods~\cite{zhou2017anomaly, lai2020robust, yu2021normality} and iterative methods~\cite{xia2015learning, beggel2019robust, fan2020robust, pang2020self}. 
Robust methods modify model architectures or 
loss functions to handle anomalies in a dataset implicitly. 
Iterative methods explicitly select pseudo-normal and pseudo-abnormal samples and use them in an iterative learning process. 
The amounts of pseudo-labeled samples are critical hyper-parameters for iterative approaches.
However, in unsupervised AD, a validation set is often not available. Thereby, tuning hyper-parameters is not a valid option.
Therefore, they are determined manually as fixed values in the previous studies~\cite{beggel2019robust, fan2020robust, pang2020self}, 
although it is not an optimal approach as the ratio of abnormal samples in a dataset, \ie, anomaly ratio, may vary in different tasks.

In this paper, we propose an unsupervised deep OCC that learns normality from pseudo-labeled normal samples.
Unlike the previous approaches based on pseudo-labeling that rely on hyper-parameters correlated with anomaly ratio, 
our method implicitly estimates the anomaly ratio of a dataset,
thereby removing the dependency on hyper-parameters.
In unsupervised AD scenarios, a learned representation is unreliable as abnormal samples detrimentally affect the training.
Moreover, we observe that the anomaly ranking of each sample fluctuates during training even after convergence.
Inspired by the observation, we propose to exploit ranking-based training dynamics of samples to find a threshold that captures anomalies with high precision and recall.
Our key idea stems from the fundamental principle in AD, \aka, concentration assumption~\cite{ruff2021unifying}.
That is, while the data space is unbounded, normal samples lie in high-density regions, and this normality region can be bounded.
In contrast, abnormal samples need not be concentrated but lie in low-density regions.
We hypothesize that normal samples are more likely to fluctuate within the normality regions while abnormal samples fluctuate outside the normality regions during model training.
Therefore, an effective threshold can be captured by measuring 
the fluctuations across the two regions, \ie, 
within and outside of the normality region.
Then, we propose to utilize the pseudo-normal samples separated by the threshold for normality learning.
The proposed method is free from hyper-parameters for pseudo-labeling and applies to a training dataset with unknown anomaly ratios.
We evaluate the effectiveness of the proposed method on various datasets, which are 10 anomaly detection benchmarks with different levels of anomaly ratios. 
We further analyze that the pseudo-labeling by our method captures anomalies with higher precision and recall than the previous methods~\cite{xia2015learning, ruff2018deep}. 

\section{Related Work}
\noindent\textbf{One-class classification-based anomaly detection.}
One-class classification (OCC) identifies objects of a specific class amongst all objects by primarily learning from a training dataset consisting of only objects of the target class~\cite{perera2021one}.
OCC has been widely applied to anomaly detection (AD) by describing normality. 
%
The most studied methods as a solid foundation for OCC-based AD are One-Class SVM (OC-SVM)~\cite{scholkopf2001estimating} and Support Vector Data Description (SVDD)~\cite{tax2004support}.
They find a hyperplane and a hypersphere enclosing most of the training data, respectively. 
By leveraging recent successes in deep learning, many studies have integrated the concept of OC-SVM or SVDD into a deep model for AD~\cite{oza2018one, ruff2018deep,  perera2019ocgan, wu2019deep, goyal2020drocc}.
However, they commonly assume a training dataset consists of only normal samples.
In formulating AD as an OCC, this fundamental assumption debases the model's applicability in the real world.

\noindent\textbf{Unsupervised anomaly detection.}
Unsupervised AD approaches have been proposed to tackle learning normality from anomaly mixed datasets, which aim to reduce the negative learning effects of abnormal data.
Robust methods adopt the idea analogous to RPCA~\cite{candes2011robust} by convex relaxation~\cite{zhou2017anomaly} or projecting a latent feature into a lower-dimensional subspace to detect anomalies.
NCAE~\cite{yu2021normality} uses adversarial learning to calibrate a latent distribution robust to outliers. 
Although they do not explicitly use pseudo-labeling, loss and regularizer weighting play an analogous role. 
On the other hand, iterative methods~\cite{beggel2019robust,  xia2015learning, fan2020robust, pang2020self} explicitly define pseudo-normal and pseudo-abnormal samples and use them in an iterative learning process. However, because the optimal ratio of pseudo-normal and pseudo-abnormal samples are unknown, most of the previous works~\cite{beggel2019robust, fan2020robust, pang2020self} use fixed ratios when selecting pseudo-normal and abnormal samples. Unlike the other iterative approaches, Xia~\etal~\cite{xia2015learning} propose to find a threshold 
by minimizing the intra-class variances of pseudo-normal and pseudo-abnormal samples. 

\noindent\textbf{Training dynamics.}
Recently, training dynamics, \ie traces of SGD or logits during training, is being used in analyzing catastrophic forgetting~\cite{toneva2018empirical}, measuring the importance of data samples on a learning task~\cite{ghorbani2019data}, large dataset analysis~\cite{swayamdipta2020dataset}, and identifying noisy labels~\cite{pleiss2020identifying}. 
Toneva \etal~\cite{toneva2018empirical} use training dynamics to identify forgetting events over the course of training.
Ghorbani \etal~\cite{ghorbani2019data} propose a method called Data Shapley to quantify the predictor performance of each training sample to identify potentially corrupted data samples. Dataset Cartography~\cite{swayamdipta2020dataset} analyzes a large-scale dataset map with training dynamics to identify the presence of ambiguous-, easy-, and hard-to-learn data regions in a feature space.
In noisy label learning~\cite{pleiss2020identifying}, training dynamics, a margin between the top two logit values during training, is used to identify mislabeled samples. 
Unlike previous approaches using SGD or logits for training dynamics, we introduce new ranking-based training dynamics 
and apply them to unsupervised AD. 

\section{Preliminary}
We provide a brief introduction to Deep SVDD~\cite{ruff2018deep}, which is a prevailing one-class classification-based deep anomaly detection model.
We utilize Deep SVDD as a base model in this paper.
Given a training dataset $\calD=\{\boldsymbol x_1, \cdots, \boldsymbol x_n\}$, where $\boldsymbol x_i \in \mathbb{R}^d$,
Deep SVDD maps data $\boldsymbol x\in\calX$ to a feature space $\calF$ through $\phi(\cdot;\calW): \calX \rightarrow \calF$, while gathering data around the center $\boldsymbol c$ of a hypersphere in the feature space, where $\calW$ denotes the set of weights of the neural network $\phi$.
Then, the sample loss $l$ for Deep SVDD is defined as follows:
\begin{equation}
    \label{eqn:dsvdd-sb}
    l(\boldsymbol x_i) = \nu \cdot R^2 + \textrm{max}(0, \norm{\phi(\boldsymbol x_i; \calW) - \boldsymbol c}^2-R^2).
\end{equation}
For simplicity, we omit regularizers in the rest of this section.
The hyper-parameter $\nu$ controls the trade-off between the radius $R$ and the amounts of data outside the hypersphere.
In the case that most of the training data are normal, the above sample loss can be simplified, which is defined as follows:
\begin{equation}
    \label{eqn:dsvdd-oc}
    l(\boldsymbol x_i) = \norm{\phi(\boldsymbol x_i; \calW) - \boldsymbol c}^2.
\end{equation}
These two versions of Deep SVDD are called \textit{soft-boundary} Deep SVDD and \textit{One-Class} Deep SVDD.
An anomaly score in both models is measured by the distance between the center and a data point (\Eref{eqn:dsvdd-score}).
\begin{equation}
    \label{eqn:dsvdd-score}
    s(\boldsymbol x_i;\calW) = \norm{\phi(\boldsymbol x_i;\calW)-\boldsymbol c}^2.
\end{equation}

\section{Unsupervised Deep One-class Classification}
In unsupervised anomaly detection (AD) scenarios, a deep one-class classification (OCC) model assuming a training dataset consisting of only normal samples would learn contaminated normality by abnormal samples mixed in a training dataset.
For better robust normality learning, we propose to add one more step, pseudo-labeling, for each training epoch.
To this end, our unsupervised deep OCC model tracks changes in the ranking of anomaly scores for each data sample, \ie, \textit{ranking-based training dynamics}, and identifies the adaptive threshold where significant ranking changes rarely happen.
Then the data samples within the adaptive threshold
are utilized as pseudo-normal samples in the following training epoch.
Our pseudo-labeling step does not require any hyper-parameter or prior knowledge of a dataset, such as an anomaly ratio.

\subsection{Ranking-based Training Dynamics and Threshold Selection}
\begin{figure}[t]
    \centering
    \includegraphics[width=0.7\columnwidth]{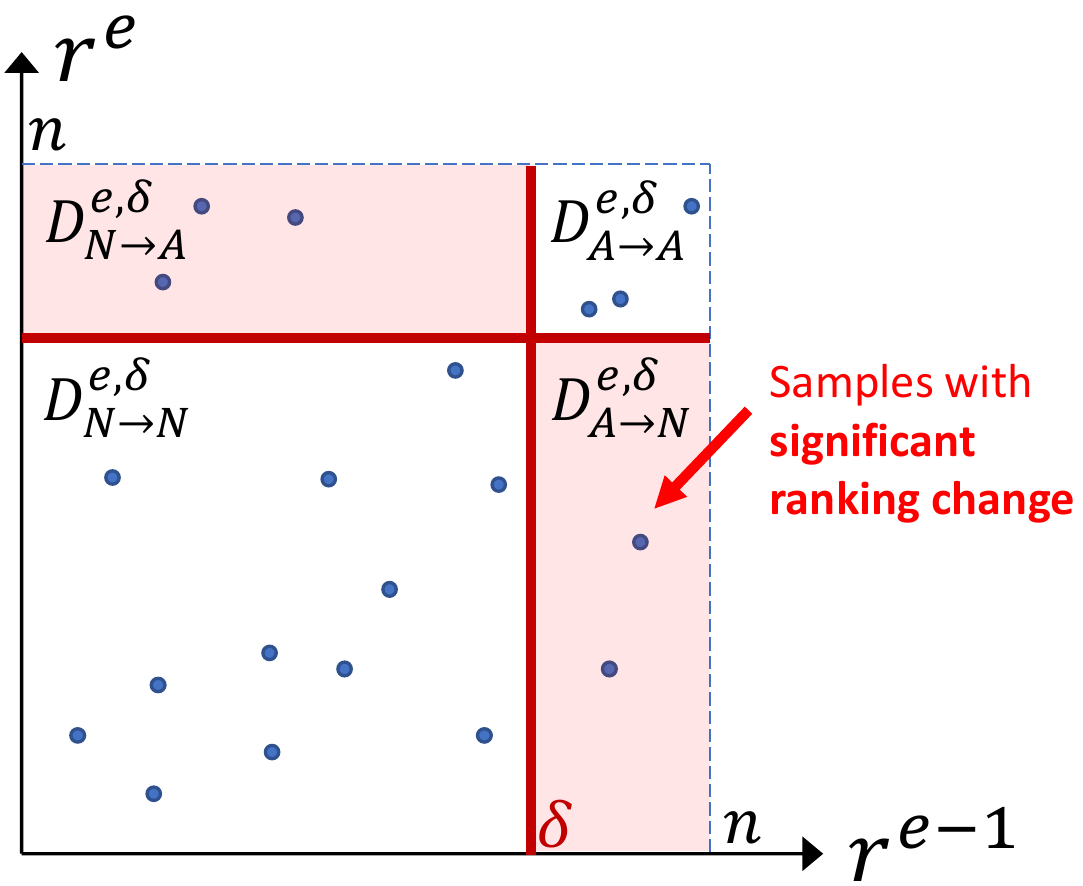}
    \caption{Concept of ranking-based training dynamics and samples with significant ranking change. 
    Dots on the two-dimensional plane 
    denote the ranking-based training dynamics of a training dataset at epoch $e$.
    Samples in the colored area denote the samples with significant ranking change according to the threshold $\delta$.}
    \label{fig:method1}
\end{figure}

We define ranking-based training dynamics for each data sample as a two-dimensional vector whose elements are rankings of anomaly scores in adjacent epochs.
We denote the anomaly score of data $\boldsymbol x_i$ at epoch $e$ as 
$s_i^e\in\mathbb{R}$,
which is measured by \Eref{eqn:dsvdd-score}, where $i\in\{1,\cdots,n\}$, $e\in\{1,\cdots,E\}$, $n$ is the number of data in a training dataset, and $E$ is the number of epochs.
Then, we obtain a sorted list of $s_i^e$ in ascending order and denote the ranking of $s_i^e$ in the sorted list as 
$r_i^e\in\mathbb{N}$,
where $r_i^e\in\{1,\cdots,n\}$.
The ranking-based training dynamics for $\boldsymbol x_i$ at epoch $e$ is defined as $(r_i^{e-1}, r_i^e)$.
The dots in \Fref{fig:method1} denote the ranking-based training dynamics of each data sample at epoch $e$.

\begin{figure}[t]
    \centering
    \includegraphics[width=0.95\columnwidth]{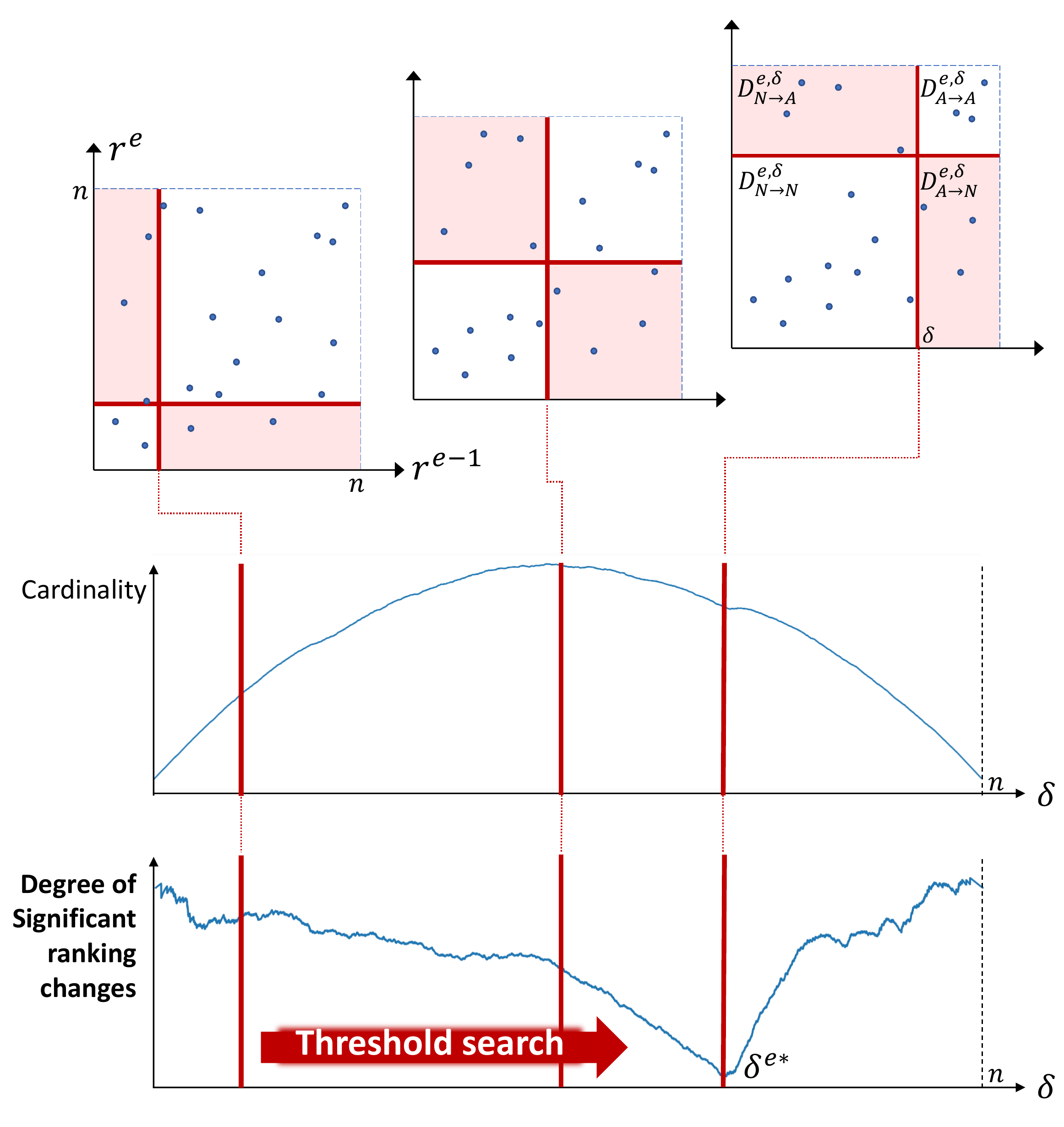}
    \caption{Concept of threshold selection.
    The degree of significant ranking changes between epoch $e-1$ and $e$ are calculated for all possible threshold $\delta$. Then, the threshold $\delta^{e*}$ with the minimum changes is used for pseudo-normal sample selection.
    (Middle) Example of the cardinality of a sample set with significant ranking changes according to threshold $\delta$.
    (Bottom) Example of the degree of significant ranking changes which is the scaled cardinality of the same sample set. 
    Both graphs are obtained
    with the Satellite dataset. 
    }
    \label{fig:method2}
\end{figure}

Starting from the second epoch, $e=2$, we adaptively select a threshold $\delta$ separating the normality region based on the ranking-based training dynamics.
To this end, we define significant ranking change motivated by the following observation and hypothesis:
When model parameters converge during training, the anomaly scores are also expected to converge, \ie, the rankings of anomaly scores are also expected to no longer change.
In practice, due to a deep model's stochastic training and high complexity, the rankings of anomaly scores frequently change over epochs, even after a sufficient number of epochs.

However, not all the ranking changes are important. 
The samples with low and high anomaly scores are likely to be normal and abnormal samples with high confidence.
Assuming we have an ideal threshold,
the local ranking changes among normal samples (low anomaly scores) or among abnormal samples (high anomaly scores) do not affect the capability of the threshold separating the normality region.
Significant ranking changes occur when rankings below the threshold, \ie, samples in the normality region, change to rankings over the threshold, \ie, outside the normality region, or vice versa.
Our hypothesis is that significant ranking changes based on the ideal threshold are less likely to occur.

We divide the two-dimensional space in which ranking-based training dynamics exist into four regions to examine significant ranking changes according to a threshold $\delta$ (\Eref{eqn:sets}).
Examples of four regions are shown in \Fref{fig:method1}.
$N$ and $A$, which are the subscripts of \Eref{eqn:sets}, denote a set of pseudo-normal and pseudo-abnormal
samples, respectively.
\begin{equation*}
    N = \{\boldsymbol x_i | r_i < \delta\}, \quad A = \{\boldsymbol x_i | r_i \geq \delta\}
\end{equation*}
A set of samples with significant ranking changes based on threshold $\delta$ is defined as the union of $\calD_{N\rightarrow A}^{e,\delta}$ and $\calD_{A\rightarrow N}^{e,\delta}$ (areas colored red in \Fref{fig:method1}).
Finally, the degree of significant ranking changes is calculated for all possible $\delta\in\{2,\cdots n\}$, as in Otsu's method~\cite{otsu1979threshold}, and the $\delta$ with the minimum degree of significant ranking changes is set as the adaptive threshold $\delta^{e*}$ at epoch $e$ (\Eref{eqn:ab}).
One important thing to note is that
since the cardinality of a set $\calD_{N\rightarrow A}^{e,\delta} \cup \calD_{A\rightarrow N}^{e,\delta}$ is naturally proportional to the area corresponding to sets in a two-dimensional space,
the minimum significant ranking changes occur 
when a threshold is close to both ends
regardless of tasks without proper scaling.
To deal with this problem, we scale the 
cardinality by the corresponding area $\delta\times(n-\delta)$ as in \Eref{eqn:ab}.
The example of threshold selection is shown in \Fref{fig:method2}.
Then, the pseudo-normal set $\calD_N^{e+1}$ for model training in the next epoch $e+1$ is determined as in \Eref{eqn:pn}.
For better stability during training, all threshold values are averaged before performing pseudo-labeling.

\begin{align}
\begin{split}
    \calD_{N\rightarrow N}^{e,\delta} &= \{ \boldsymbol x_i \ | \ r_i^{e-1} < \delta, \ r_i^e < \delta \}, \\
    \calD_{N\rightarrow A}^{e,\delta} &= \{ \boldsymbol x_i \ | \ r_i^{e-1} < \delta, \ r_i^e \geq \delta \}, \\
    \calD_{A\rightarrow N}^{e,\delta} &= \{ \boldsymbol x_i \ | \ r_i^{e-1} \geq \delta, \ r_i^e < \delta \}, \\
    \calD_{A\rightarrow A}^{e,\delta} &= \{ \boldsymbol x_i \ | \ r_i^{e-1} \geq \delta, \ r_i^e \geq \delta \},
    \label{eqn:sets}
\end{split}
\end{align}

\begin{equation}
    \label{eqn:ab}
    \delta^{e*} = \argmin_\delta \frac{1}{\delta\times(n-\delta)} \ (|\calD_{N\rightarrow A}^{e,\delta}| + |\calD_{A\rightarrow N}^{e,\delta}|),
\end{equation}

\begin{equation}
\begin{split}
    \label{eqn:pn}
    \calD_N^{e+1} = \{\boldsymbol x_i | r_i^{e-1} < \bar{\delta}^{e*}, r_i^e < \bar{\delta}^{e*} \}, \quad
    \bar{\delta}^{e*} = \frac{1}{|\Delta^e|}\sum_{\delta \in \Delta^e}\delta, \\
    \text{where} \quad \Delta^e = \{\delta^{e'*} | e' \leq e\}.
\end{split}
\end{equation}

\begin{algorithm}[t]
\caption{Unsupervised Deep OCC with training dynamics}
\label{alg:algorithm}
\textbf{Input}: Unlabeled dataset $\calD=\{\boldsymbol x_1, \cdots, \boldsymbol x_n\}$, 
\newline \hspace*{9mm} Maximum number of epochs $E$ \\
\textbf{Output}: Anomaly scores, Threshold

\begin{algorithmic}[1] 
\STATE Initialize the training set: $\calD_N^1 = \calD_N^2 = \calD$
\FOR{$e=1$ to $E$}
\STATE Train model with $\calD_N^e$ and $\calL^e$ (\Eref{eqn:oc-u})
\IF{$e\ge2$}
\STATE Calculate ranking-based training dynamics $(r_i^{e-1}, r_i^e)$\\
\STATE Search threshold $\delta^{e*}$ by ranking changes (\Eref{eqn:ab})
\STATE Update the pseudo-normal set $\calD_N^{e+1}$ (\Eref{eqn:pn})
\ENDIF
\ENDFOR
\end{algorithmic}
\end{algorithm}

\subsection{Unsupervised Deep One-class Classification}
Our proposed unsupervised deep one-class classification utilizes the pseudo-normal samples updated by ranking-based training dynamics every epoch, except the first two epochs where all the training data are used.
The total loss at epoch $e$, $\calL^e$, is defined as follows:

\begin{equation}
\label{eqn:oc-u}
  \setlength{\arraycolsep}{0pt}
  \calL^e = \left\{ \begin{array}{ l l }
    & \quad\, \frac{1}{n} \sum_{\boldsymbol x_i \in \calD} \norm{\phi(\boldsymbol x_i; \calW) - \boldsymbol c}^2, \quad e\leq2. \\[10pt]
    & \frac{1}{|\calD_N^e|} \sum_{\boldsymbol x_i \in \calD_N^e} \norm{\phi(\boldsymbol x_i; \calW) - \boldsymbol c}^2, \quad e\geq3.
  \end{array} \right.
\end{equation}

We outline the learning process of unsupervised deep one-class classification in \Aref{alg:algorithm}.

\section{Experiment}

\subsection{Experiment Settings}
\subsubsection{Datasets and evaluation metrics}\label{sec:DatasetnMetric}
\textit{Anomaly detection benchmarks}~\cite{rayana2016odds}
consist of multivariate tabular datasets. 
As shown in \Tref{tab:dataset}, each dataset has $n$ number of data samples with $d$ number of attributes.
These benchmarks have a wide range of anomaly ratios in a dataset, from 1.2\% to 34.9\%.

We use three performance metrics for binary classifiers as evaluation metrics in this paper: the area under the receiver operating characteristic curve (ROCAUC), the area under the precision-recall curve (PRAUC), and F1-Score.
ROCAUC and PRAUC are calculated on the predicted anomaly scores by varying a decision threshold.
ROCAUC is known to be over-confident when the classes are highly imbalanced, which is often the case for anomaly detection, where the anomaly ratio is usually low. 
Therefore, we also report PRAUC, which scales performance according to anomaly ratio.
F1-Score measures the discriminativeness of the selected thresholds.

\begin{table}[t]
\caption{Anomaly detection benchmarks.}
\begin{center}
\begin{tabular}{|c|c|c|c|c|}
\hline
Dataset     & Points    & Dim.  & Anomalies & Anomaly ratio (\%)\\ \hline
Pima        &  768      &   8   &  268      & 34.9          \\ \hline
Satellite   & 6435      &  36   & 2036      & 31.6          \\ \hline
Arrhythmia  &  452      & 274   &   66      & 14.6          \\ \hline
Cardio      & 1831      &  21   &  176      &  9.6          \\ \hline
Mnist       & 7603      & 100   &  700      &  9.2          \\ \hline
Wbc         &  378      &  30   &   21      &  5.6          \\ \hline
Glass       &  214      &   9   &    9      &  4.2          \\ \hline
Thyroid     & 3772      &   6   &   93      &  2.5          \\ \hline
Pendigits   & 6870      &  16   &  156      &  2.3          \\ \hline
Satimage-2  & 5803      &  36   &   71      &  1.2          \\ \hline
\end{tabular} \label{tab:dataset}
\end{center}
\end{table}

\subsubsection{Competing methods}
The proposed method is compared with the following deep one-class classification methods:
\begin{itemize}
    \item \textit{soft-boundary} Deep SVDD~\cite{ruff2018deep}:
    A deep one-class classifier that explicitly learns a decision boundary while enforcing a fraction of data to lie outside of the boundary through a hyper-parameter $\nu$. For the hyper-parameter, we use a true anomaly ratio according to $\nu$-property~\cite{scholkopf2001estimating, ruff2018deep}.
    \item \textit{One-Class} Deep SVDD~\cite{ruff2018deep}:
    A deep one-class classifier designed under the assumption that all the training data is normal.
    \item \textit{One-Class} Deep SVDD + Otsu's method:
    \textit{One-Class} Deep SVDD trained with pseudo-normal samples selected by a threshold. Otsu's method~\cite{otsu1979threshold} is applied to anomaly scores as in the study~\cite{xia2015learning} to search for the threshold at each epoch.
    Dataset is divided into two groups by the possible thresholds, and the threshold that minimizes the intra-class variance is selected.
    \item \textit{One-Class} Deep SVDD + Pre-defined threshold:
    \textit{One-Class} Deep SVDD trained with pseudo-normal samples selected according to a pre-defined threshold. We use each dataset's true anomaly ratio (TAR) as the threshold.
\end{itemize}
We denote each competing method as SB, OC, OC+Otsu, and OC+TAR, respectively.

\begin{table*}[t!]
\caption{
Performance results measured by ROCAUC and PRAUC.
We report the average AUC with a standard deviation computed over 10 seeds.
The highest performances among OCC-based deep AD models are indicated in bold.
}
\begin{center}
{\small 
\begin{tabular}{|c|c|c|c|c|c|c|c|c|c|c|}
\hline
\multirow{2}{*}{Dataset} & \multicolumn{4}{c|}{ROCAUC}  & \multicolumn{4}{c|}{PRAUC}    \\ \cline{2-9} 
                         & SB               & OC                & OC+Otsu                   & Proposed                  & SB                & OC                & OC+Otsu           & Proposed                 \\ \hline
Pima                     & $54.9 \pm 3.8$   & $56.6 \pm 4.7$    & $57.1 \pm 5.1$            & $\mathbf{58.6 \pm 4.7}$   & $38.7 \pm 2.6$    & $39.7 \pm 3.5$    & $40.5 \pm 3.8$    & $\mathbf{42.4 \pm 3.7}$  \\ 
Satellite                & $65.6 \pm 5.5$   & $68.7 \pm 4.4$    & $71.8 \pm 5.6$            & $\mathbf{77.6 \pm 3.7}$   & $58.6 \pm 7.2$    & $63.3 \pm 6.6$    & $67.6 \pm 6.1$    & $\mathbf{76.0 \pm 3.0}$  \\ 
Arrhythmia               & $69.6 \pm 4.0$   & $68.1 \pm 3.6$    & $70.4 \pm 3.2$            & $\mathbf{71.4 \pm 3.3}$   & $26.4 \pm 3.1$    & $27.6 \pm 3.8$    & $33.2 \pm 2.9$    & $\mathbf{35.3 \pm 2.9}$  \\ 
Cardio                   & $72.4 \pm 6.2$   & $80.6 \pm 4.4$    & $82.0 \pm 4.0$            & $\mathbf{85.2 \pm 4.5}$   & $23.6 \pm 5.9$    & $31.8 \pm 7.2$    & $35.7 \pm 6.6$    & $\mathbf{44.5 \pm12.0}$  \\ 
Mnist                    & $69.8 \pm 3.4$   & $77.4 \pm 4.4$    & $77.4 \pm 5.4$            & $\mathbf{79.4 \pm 4.7}$   & $21.2 \pm 2.2$    & $28.7 \pm 3.3$    & $29.8 \pm 4.5$    & $\mathbf{32.5 \pm 4.2}$  \\ 
Wbc                      & $75.2 \pm 6.4$   & $82.5 \pm 6.6$    & $82.6 \pm 5.0$            & $\mathbf{85.4 \pm 3.1}$   & $20.2 \pm10.2$    & $26.7 \pm12.9$    & $33.3 \pm10.9$    & $\mathbf{34.5 \pm11.3}$  \\ 
Glass                    & $66.3 \pm 8.8$   & $77.9 \pm 9.3$    & $\mathbf{79.2 \pm10.1}$   & $\mathbf{79.2 \pm 9.6}$   & $12.0 \pm 7.0$    & $14.5 \pm 6.0$    & $14.2 \pm 6.6$    & $\mathbf{16.2 \pm 6.8}$  \\ 
Thyroid                  & $78.2 \pm 5.3$   & $90.9 \pm 4.0$    & $90.7 \pm 3.3$            & $\mathbf{91.4 \pm 3.7}$   & $14.7 \pm 5.4$    & $25.3 \pm 6.9$    & $27.9 \pm 9.7$    & $\mathbf{36.0 \pm11.7}$  \\ 
Pendigits                & $64.6 \pm11.0$   & $71.1 \pm 8.5$    & $75.5 \pm 7.3$            & $\mathbf{76.8 \pm10.4}$   & $ 5.3 \pm 1.4$    & $ 6.2 \pm 2.6$    & $ 8.7 \pm 4.4$    & $\mathbf{13.6 \pm13.5}$  \\ 
Satimage-2               & $77.1 \pm 8.2$   & $93.0 \pm 3.9$    & $94.4 \pm 3.1$            & $\mathbf{95.1 \pm 2.4}$   & $ 8.3 \pm 4.0$    & $15.1 \pm 6.7$    & $19.5 \pm 9.3$    & $\mathbf{22.3 \pm15.9}$  \\ \hline
\end{tabular}
}
\label{tab:auc}
\end{center}
\end{table*}

\begin{table}[ht]
\caption{
Performance results measured by F1-Score.
We report the average score with a standard deviation computed over 10 seeds.
The highest performances are indicated in bold.
}
\begin{center}
{\small
\begin{tabular}{|c|c|c|c|c|}
\hline
\multirow{2}{*}{Dataset} & \multicolumn{3}{c|}{F1-Score}    \\ \cline{2-4} 
                         & SB                       & OC+Otsu           & Proposed                  \\ \hline
Pima                     & $\mathbf{38.4 \pm 3.9}$  & $ 5.5 \pm 2.9$    & $18.2 \pm 5.7$            \\ 
Satellite                & $50.8 \pm 5.4$           & $ 5.8 \pm 5.2$    & $\mathbf{65.7 \pm 3.4}$   \\ 
Arrhythmia               & $29.0 \pm 5.9$           & $17.6 \pm 4.3$    & $\mathbf{30.1 \pm 8.4}$   \\ 
Cardio                   & $24.1 \pm 6.7$           & $ 8.8 \pm 6.6$    & $\mathbf{44.8 \pm 9.6}$   \\ 
Mnist                    & $26.2 \pm 4.4$           & $ 4.4 \pm 3.2$    & $\mathbf{33.8 \pm 2.8}$   \\ 
Wbc                      & $23.5 \pm11.9$           & $27.1 \pm 8.0$    & $\mathbf{32.7 \pm11.4}$   \\ 
Glass                    & $\mathbf{17.5 \pm 7.5}$  & $16.1 \pm 3.8$    & $13.8 \pm 2.8$            \\ 
Thyroid                  & $22.8 \pm 6.7$           & $ 8.4 \pm 3.6$    & $\mathbf{36.7 \pm 9.7}$   \\ 
Pendigits                & $ 7.3 \pm 2.9$           & $ 8.3 \pm10.2$    & $\mathbf{16.3 \pm16.3}$   \\ 
Satimage-2               & $13.6 \pm 6.0$           & $14.3 \pm10.2$    & $\mathbf{17.8 \pm18.0}$   \\ \hline
\end{tabular}
}
\label{tab:f1}
\end{center}
\end{table}

\subsubsection{Implementation details}
To implement the base model, Deep SVDD\footnote{\url{https://github.com/lukasruff/Deep-SVDD-PyTorch}}, we use the source code released by the authors and adjust the backbone architectures for each dataset.
A 3-layer MLP with 128-64-32 units is used on the \textit{Arrhythmia} dataset; 
a 3-layer MLP with 64-32-16 units is used on the \textit{Mnist} dataset;
a 3-layer MLP with 32-16-4 units is used on the \textit{Pima} and \textit{Thyroid} dataset;
a 3-layer MLP with 32-16-8 units is used on the remaining 6 datasets.
The model is pre-trained with a reconstruction loss with an autoencoder for 100 epochs, and then the pre-trained encoder is fine-tuned with an anomaly detection loss for 50 epochs, \ie, $E$ is set to 50.
For an autoencoder, we utilize the aforementioned architectures for an encoder network and implement a decoder network symmetrically. 
We use Adam optimizer~\cite{kingma2015adam} with a batch size of 128 with a learning rate of $10^{-3}$. 
Data samples are standardized to have zero mean and unit variance. 
The experiments are performed with Intel Xeon Silver 4210 CPU and GeForce GTX 1080Ti GPU.

\subsection{Experiment Results}

\subsubsection{Performance}
We report unsupervised anomaly detection accuracy measured by ROCAUC and PRAUC in \Tref{tab:auc}.

When na\"ive OC is compared with OC+Otsu and our method utilizing pseudo-labeling, performance improvement is observed in most cases.
Moreover, the proposed method shows the most favorable performance.
The selected pseudo-normal samples by the proposed threshold enable OC to learn more robust normality.
We conjecture that even if Otsu's method finds the threshold that maximizes the separability between anomaly scores of two groups divided by the threshold, the threshold may not represent the boundary dividing the normal and abnormal data samples in an unsupervised AD scenario.

ROCAUC over-confidently measures performance on a dataset with a small anomaly ratio, \ie, a heavily imbalanced dataset.
Therefore, the performance improvement of anomaly detection cannot be guaranteed by only the improvement in ROCAUC for a dataset with low anomaly ratios. 
However, as aforementioned, our method also shows performance improvement in PRAUC, which considers precision adjusted by an anomaly ratio.
In addition, the performance improvement of our method 
is more pronounced in PRAUC.
These results show that our adaptive threshold applied to unsupervised one-class classification 
improves the robustness of normality on datasets with various anomaly ratios.

\begin{figure*}[t]
    \centering
    \subfigure[SB]
    {
        \includegraphics[width=2.3in]{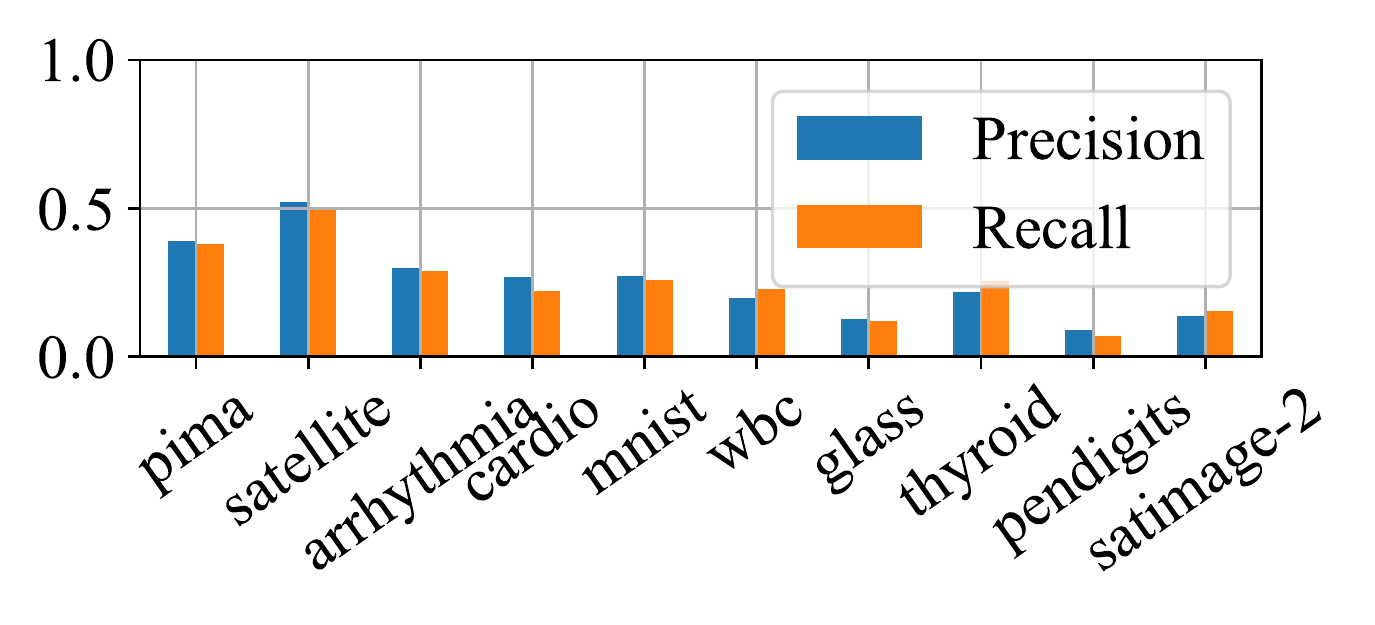}
    }
    \subfigure[OC+Otsu]
    {
        \includegraphics[width=2.2in]{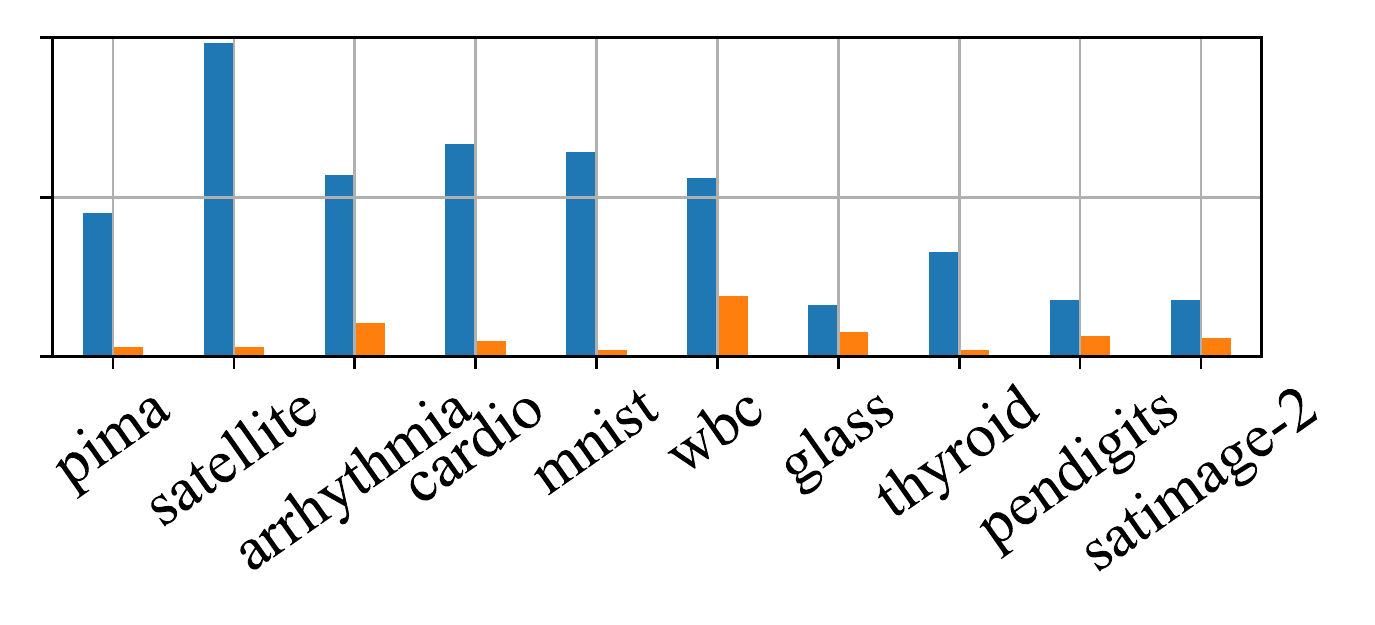}
    }
    \subfigure[Proposed]
    {
        \includegraphics[width=2.2in]{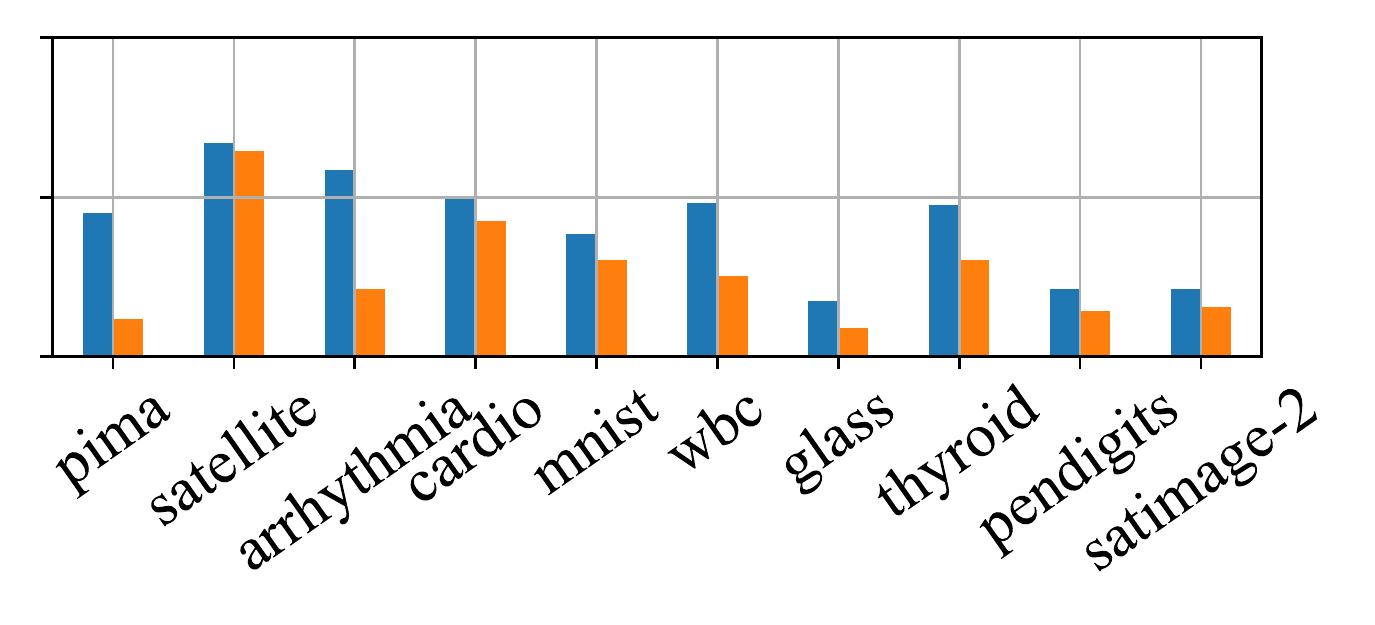}
    }
    \caption{
    Average precision (blue bar) and recall (orange bar) computed over 10 seeds in each method and each dataset. We omit standard deviations for better visibility.
    }
    \label{fig:pr_all}
\end{figure*}

\begin{figure*}[t]
    \centering
    \subfigure[ROCAUC]
    {
        \includegraphics[width=2.2in]{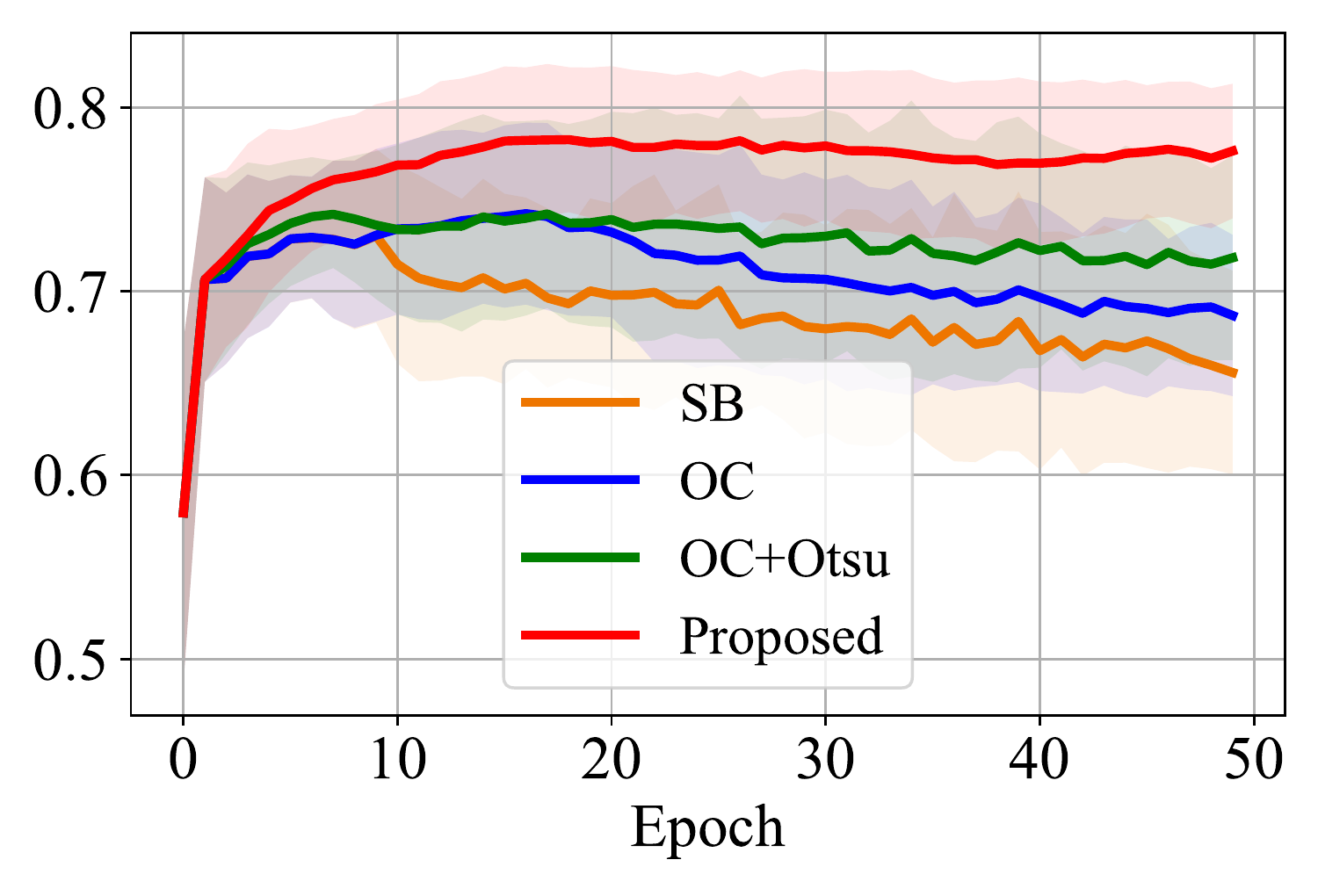}
    }
    \subfigure[PRAUC]
    {
        \includegraphics[width=2.2in]{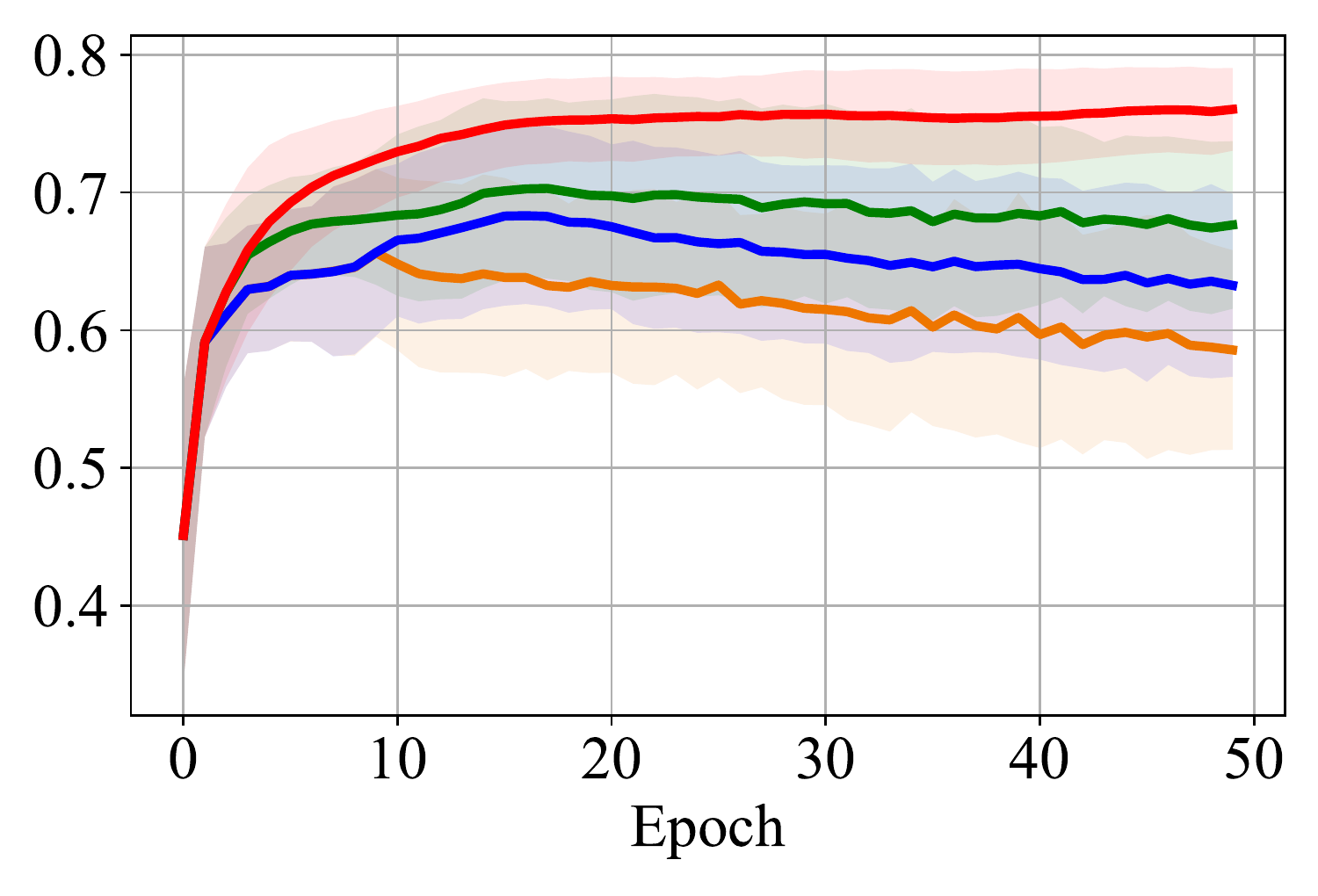}
    }
    \subfigure[F1-Score]
    {
        \includegraphics[width=2.2in]{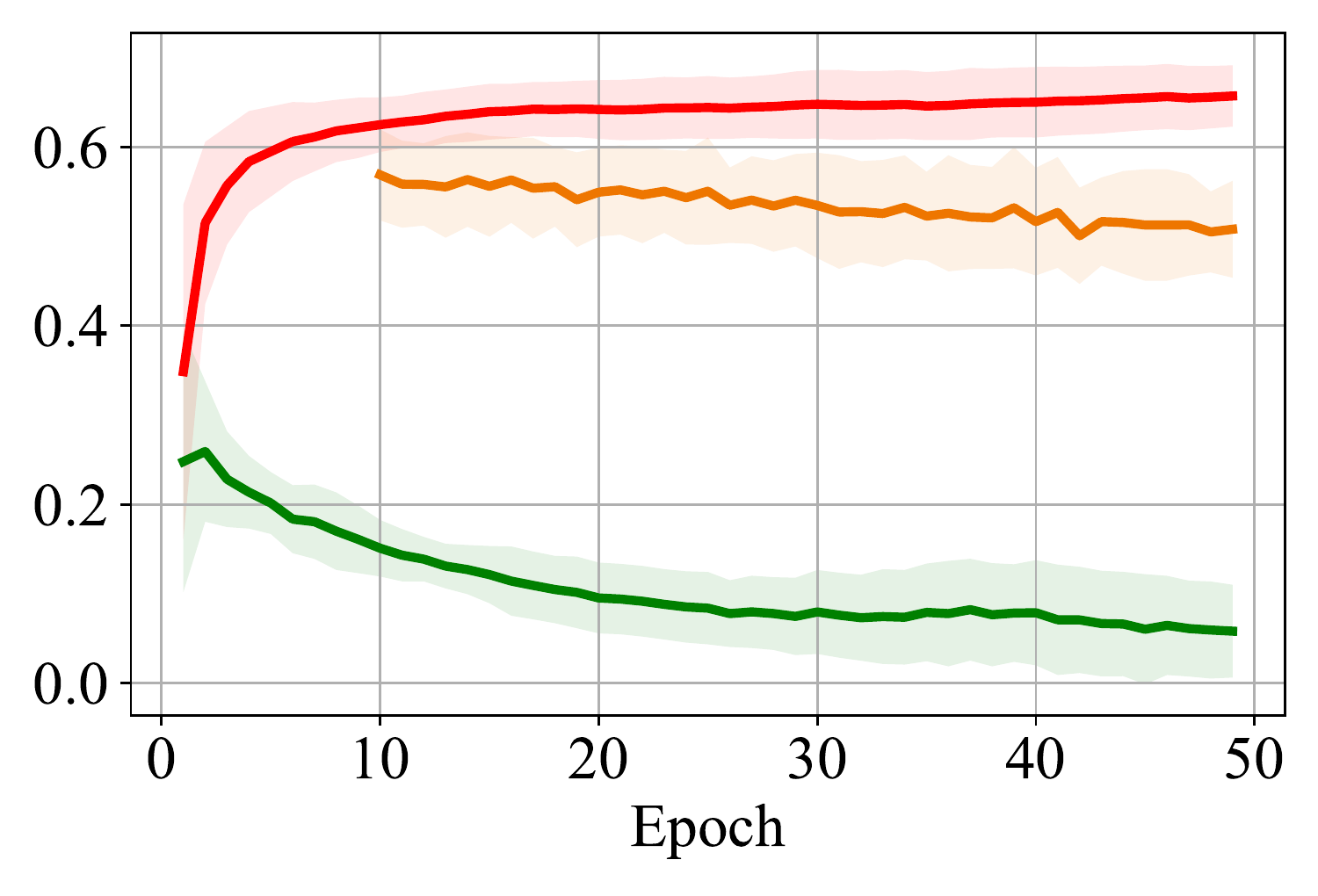}
    }
    \caption{
    Changes in (a) ROCAUC, (b) PRAUC, and (c) F1-Score measured at each epoch during model training with Satellite dataset.
    The average scores computed over 10 seeds of each evaluation metric are shown with a standard deviation represented as a shaded area.
    For SB, the model is trained with the same loss function of OC for 10 training epochs before radius R gets updated, following the released source code.
    }
    \label{fig:performance-both}
\end{figure*}  

\begin{table*}[t!]
\caption{
Performance comparison with OC trained with pseudo-normal samples selected according to a true anomaly ratio (TAR).
We report the average performance with the standard deviation computed over 10 seeds.
}
\begin{center}
{\small 
\begin{tabular}{|c|c|c|c|c|c|c|}
\hline
\multirow{2}{*}{Dataset} & \multicolumn{2}{c|}{ROCAUC}  & \multicolumn{2}{c|}{PRAUC}  & \multicolumn{2}{c|}{F1-Score}    \\ \cline{2-7} 
                         & Proposed         & OC+TAR            & Proposed          & OC+TAR            & Proposed          & OC+TAR            \\ \hline
Pima                     & $58.6 \pm 4.7$   & $61.5 \pm 5.1$    & $42.4 \pm 3.7$    & $44.0 \pm 3.9$    & $18.2 \pm 5.7$    & $45.4 \pm 4.7$    \\ 
Satellite                & $77.6 \pm 3.7$   & $77.4 \pm 3.6$    & $76.0 \pm 3.0$    & $76.0 \pm 2.9$    & $65.7 \pm 3.4$    & $65.1 \pm 3.2$    \\ 
Arrhythmia               & $71.4 \pm 3.3$   & $74.2 \pm 2.7$    & $35.3 \pm 2.9$    & $39.8 \pm 3.6$    & $30.1 \pm 8.4$    & $43.9 \pm 3.5$    \\ 
Cardio                   & $85.2 \pm 4.5$   & $85.6 \pm 4.3$    & $44.5 \pm12.0$    & $44.4 \pm11.5$    & $44.8 \pm 9.6$    & $46.0 \pm 8.7$    \\ 
Mnist                    & $79.4 \pm 4.7$   & $79.2 \pm 4.9$    & $32.5 \pm 4.2$    & $32.1 \pm 4.1$    & $33.8 \pm 2.8$    & $35.3 \pm 3.1$    \\ 
Wbc                      & $85.4 \pm 3.1$   & $86.0 \pm 3.6$    & $34.5 \pm11.3$    & $37.4 \pm14.0$    & $32.7 \pm11.4$    & $40.0 \pm13.0$    \\ 
Glass                    & $79.2 \pm 9.6$   & $79.6 \pm10.8$    & $16.2 \pm 6.8$    & $14.8 \pm 5.8$    & $13.8 \pm 2.8$    & $14.3 \pm 5.0$    \\ 
Thyroid                  & $91.4 \pm 3.7$   & $91.8 \pm 3.8$    & $36.0 \pm11.7$    & $40.2 \pm12.3$    & $36.7 \pm 9.7$    & $44.2 \pm10.3$    \\ 
Pendigits                & $76.8 \pm10.4$   & $78.2 \pm 9.1$    & $13.6 \pm13.5$    & $15.5 \pm16.8$    & $16.3 \pm16.3$    & $21.2 \pm17.4$    \\ 
Satimage-2               & $95.1 \pm 2.4$   & $94.9 \pm 2.5$    & $22.3 \pm15.9$    & $23.6 \pm17.9$    & $17.8 \pm18.0$    & $21.8 \pm20.8$    \\ \hline
\end{tabular}
}
\label{tab:tar}
\end{center}
\end{table*}           

Furthermore, we evaluate a threshold by using F1-Score (\Tref{tab:f1}).
For each competing method, a training dataset is classified by a decision boundary explicitly learned in SB and by the selected threshold in OC+Otsu and our method.
The threshold selected by our method also shows robust performance in F1-Score in most cases.
On the other hand, F1-Scores in OC+Otsu are remarkably low despite the high performance in ROCAUC and PRAUC.
To clarify the differences in classification performance, we compare the precision and recall 
of each method and dataset (\Fref{fig:pr_all}).
It is notable that OC+Otsu shows high precision but low recall, leading to a low F1-Score.
Based on these results and the concept of Otsu's method, we conjecture that the threshold selected in OC+Otsu is located far outside the normality region, and it is caused by a few abnormal samples with extremely high anomaly scores. 
It is because the output of OCC is not normalized, and an upper bound of anomaly scores does not exist. 
Proper filtering may alleviate this problem, but it leads to the problem of setting thresholds.
On the other hand, as a quantization of anomaly scores, ranking is not affected by the scale of anomaly scores.
For SB, precision and recall are almost similar. In particular, in the Pima and Glass data, F1-Score is higher than the proposed method.
Lastly, we show the changes in three evaluation metrics during model training (\Fref{fig:performance-both}).
In \Fref{fig:performance-both}, the proposed method improves and maintains the performance compared to others where the performance degrades due to the contaminated normality by abnormal samples.

\begin{figure}[ht]
    \centering
    \subfigure[Satellite (\# of data: 6435, anomaly ratio: 31.6\%)]
    {
        \includegraphics[width=3.3in]{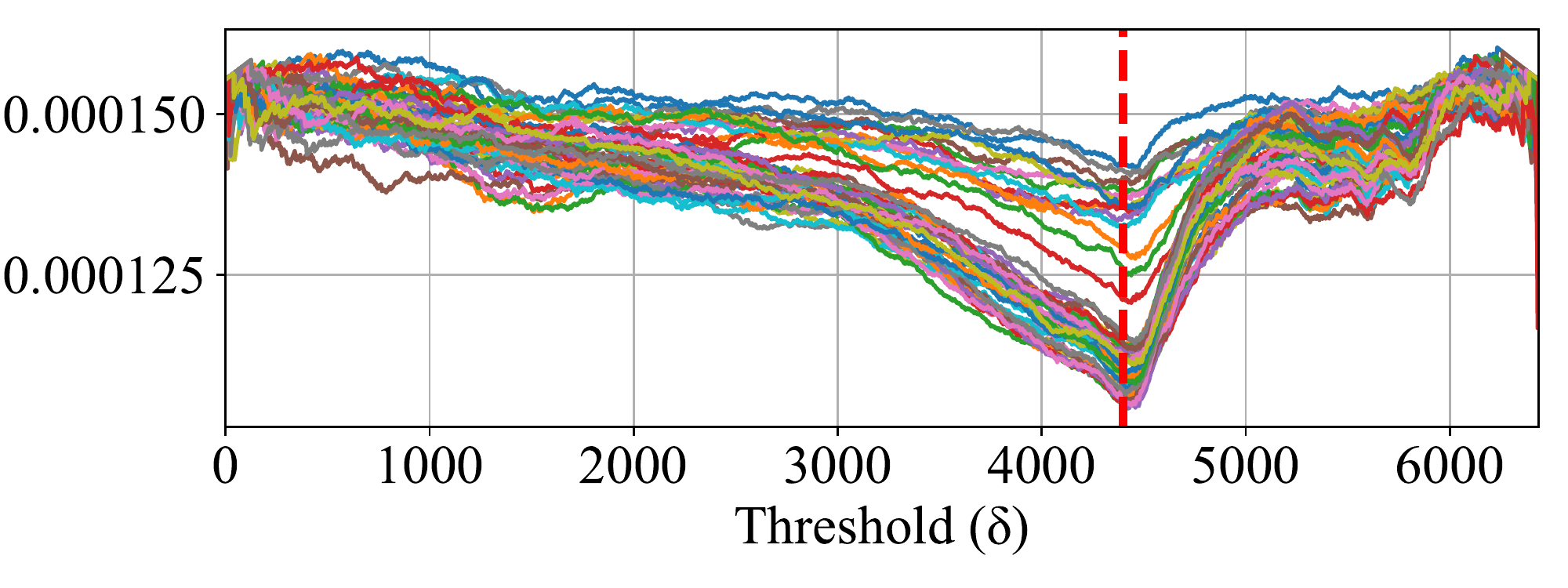}
    }
    \subfigure[Cardio (\# of data: 1831, anomaly ratio: 9.6\%))]
    {
        \includegraphics[width=3.2in]{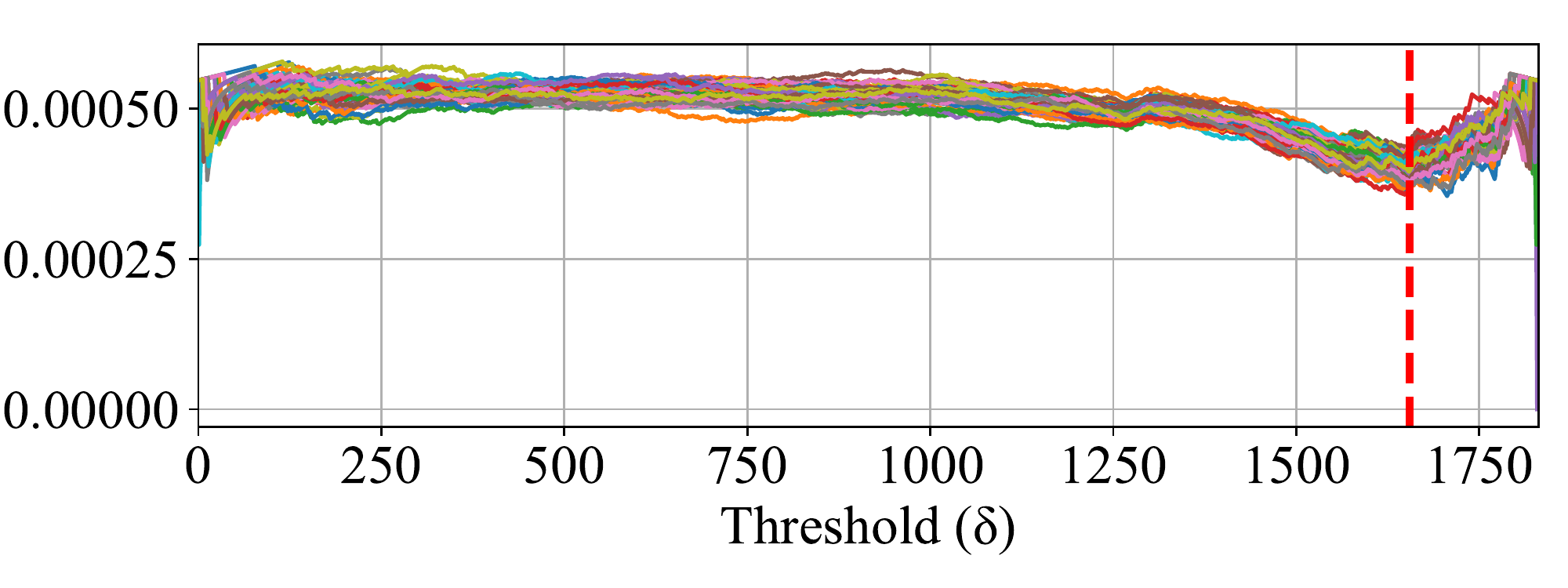}
    }
    \subfigure[Thyroid (\# of data: 3772, anomaly ratio: 2.5\%))]
    {
        \includegraphics[width=3.3in]{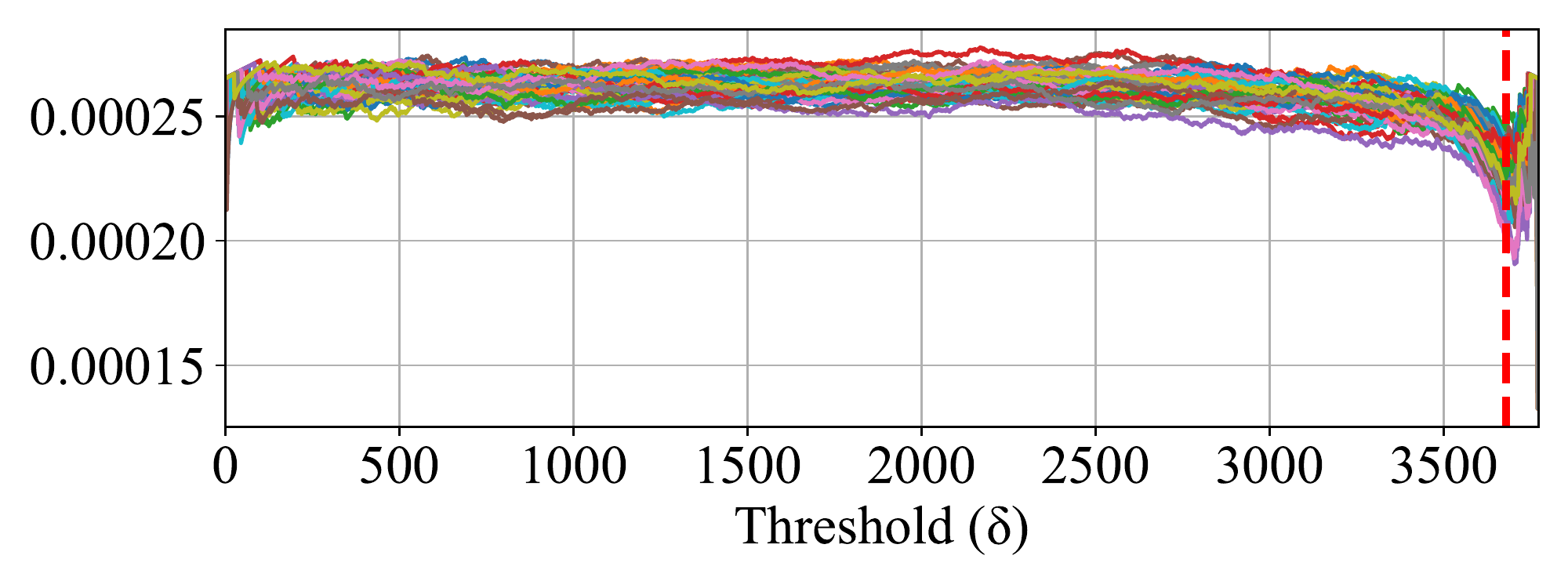}
    }
    \caption{
    The graph of the degree of significant ranking changes according to threshold $\delta$ obtained from all 50 epochs.
    The red dashed line represents the location of a true anomaly ratio in a dataset. 
    }
    \label{fig:src}
\end{figure}  
\begin{figure}[ht]
    \centering
    \subfigure[Satellite (\# of data: 6435, anomaly ratio: 31.6\%)]
    {
        \includegraphics[width=3.3in]{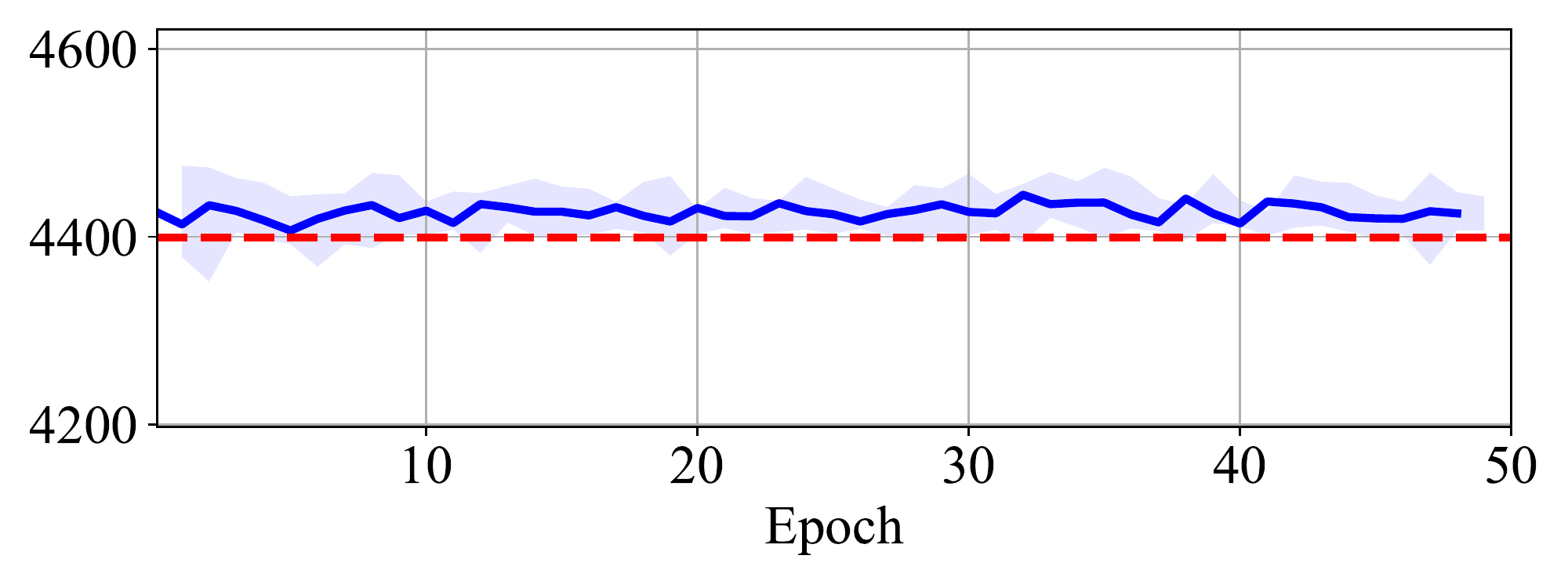}
    }
    \subfigure[Cardio (\# of data: 1831, anomaly ratio: 9.6\%))]
    {  
        \includegraphics[width=3.2in]{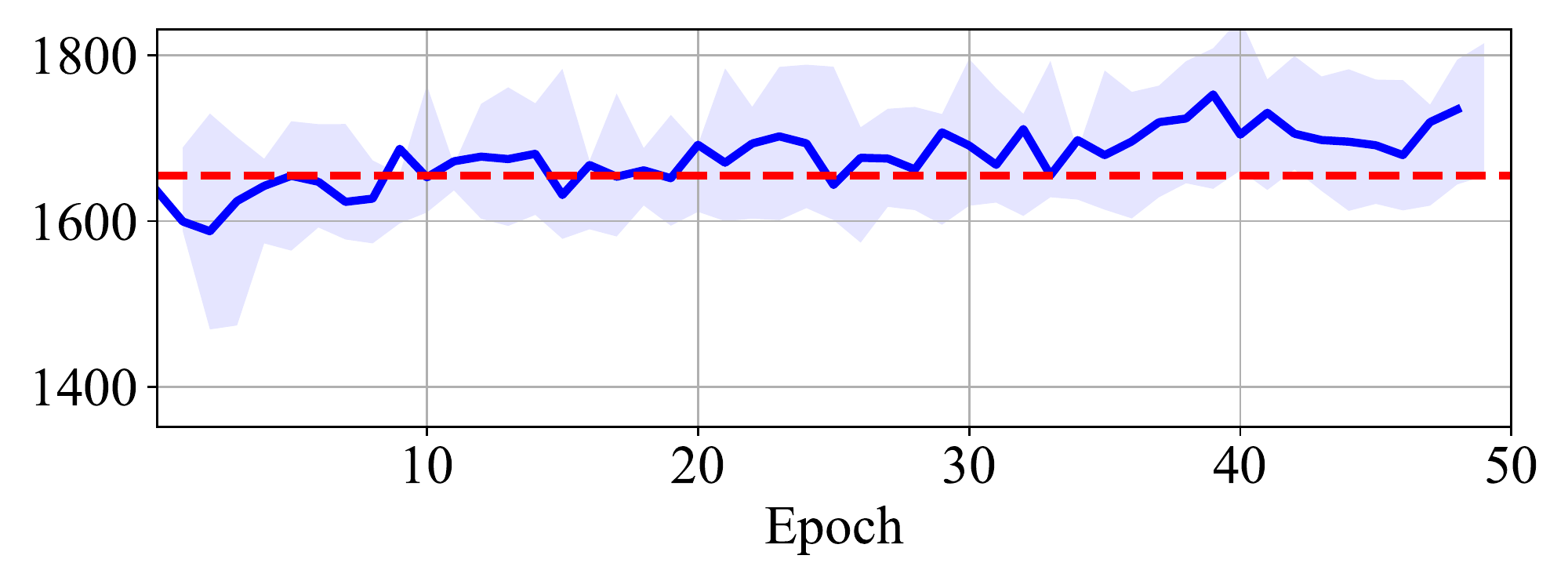}
    }
    \subfigure[Thyroid (\# of data: 3772, anomaly ratio: 2.5\%))]
    {
        \includegraphics[width=3.3in]{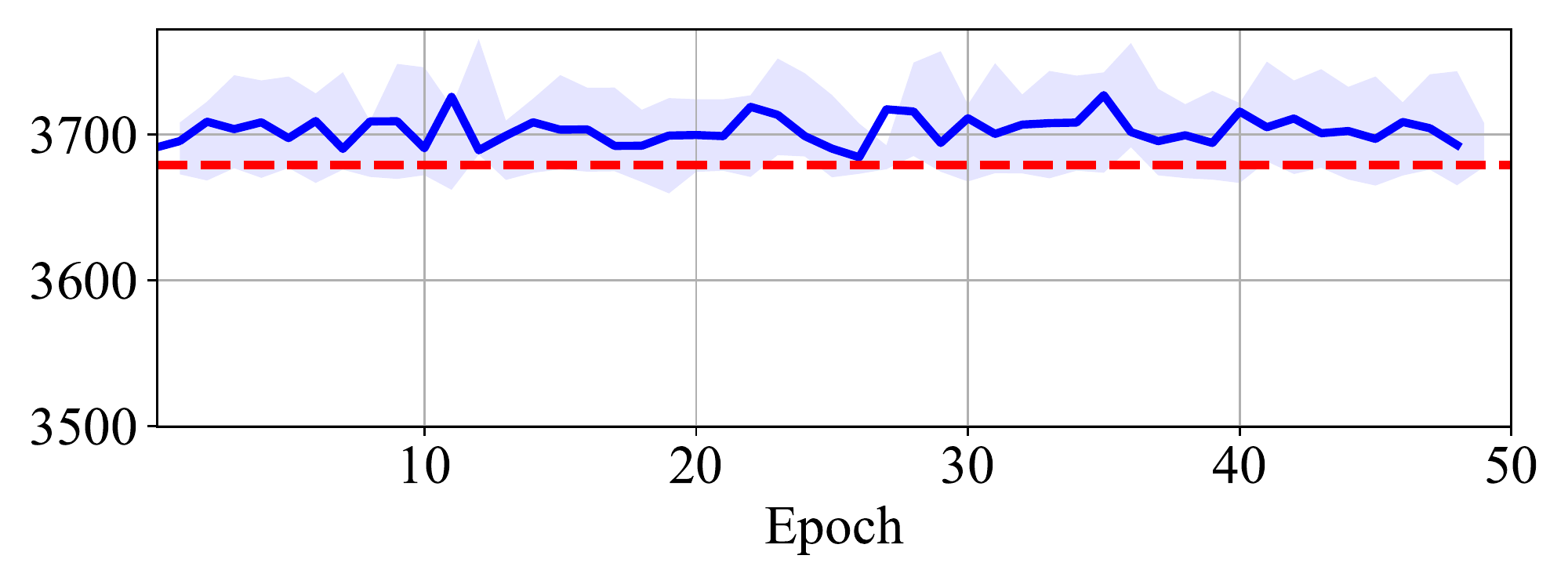}
    }
    \caption{
    The average and standard deviation of a selected threshold at each epoch during model training are shown in blue (computed over 10 seeds). 
    The red dashed line represents the location of a true anomaly ratio in a dataset.
    }
    \label{fig:thr}
\end{figure}  

\subsubsection{Comparison with oracle}
We compare our method with OC+TAR, which uses a true anomaly ratio as a hyper-parameter for pseudo labeling (\Tref{tab:tar}).
In practice, since the anomaly ratio of a task is unknown, we consider OC+TAR as an oracle method.
Our method shows competitive performance in ROCAUC and PRAUC. In a few cases, \eg, Satellite, Cardio, Mnist, Glass, and Satimage-2, our method shows on par or marginally better performance in either metric than OC+TAR. 
It is possible because the true anomaly ratio is used as a hyper-parameter for model training; therefore, it does not guarantee upper-bound performance.
In F1-Score, the performance gap is larger than that of ROCAUC or PRAUC. In ROCAUC and PRAUC, the anomaly ratio is only used as a hyper-parameter for training. However, in F1-Score, the anomaly ratio is explicitly used as a threshold to evaluate, resulting in a larger performance gap.

\subsubsection{Threshold analysis}
We visualize and analyze the threshold selected by our method compared to the true anomaly ratio.
The proposed threshold selection method is designed under the hypothesis that significant ranking changes based on the ideal threshold are less likely to occur.
For a clearer illustration, we plot the degree of significant ranking changes along the possible thresholds for 50 epochs (\Fref{fig:src}). The red dotted line represents the true anomaly ratio. 
The degree of significant ranking changes is minimized around the true anomaly ratio, which aligns with our design principle.
We visualize the selected thresholds over 50 epochs to analyze how the selected threshold is close to the true anomaly ratio over the training (\Fref{fig:thr}). The graph shows that the threshold selected by our method is close to the true anomaly ratio throughout the training.

\section{Conclusion} 
In this paper, we introduce ranking-based training dynamics tracked during model training to search for an effective threshold, which formalizes the fundamental principle in anomaly detection.
The proposed threshold selection method is applied to a one-class classification-based method to tackle unsupervised AD problems.
While many previous studies utilizing pseudo-labeling rely on a hyper-parameter to control the amount of pseudo-labeled data, no hyper-parameter tuning is required in our method. 
Moreover, the threshold analysis shows that the selected threshold is close to the true anomaly ratio. 
The experiments on various datasets with different levels of anomaly ratios validate that our method effectively improves anomaly detection performance.

\section*{Acknowledgment}
This work was supported by Institute of Information \& communications Technology Planning \& Evaluation (IITP) grant funded by the Korea government (MSIT) (No.2020-0-00833, A study of 5G based Intelligent IoT Trust Enabler).

\balance
\bibliographystyle{IEEEtran}
\bibliography{mybib}

\end{document}